\documentclass{article}
\usepackage{iclr2026_conference,times}
\iclrfinalcopy
\usepackage{hyperref}
\usepackage{url}
\usepackage{booktabs}
\usepackage{amsmath}
\usepackage{graphicx}
\usepackage{multirow}
\usepackage{xcolor}
\usepackage{subcaption}

\title{Correct Chains, Wrong Answers: Dissociating Reasoning from Output in LLM Logic}

\author{
Abinav Rao \and Sujan Rachuri \and Nikhil Vemuri
}

\begin{document}

\maketitle

\begin{abstract}
LLMs can execute every step of chain-of-thought reasoning correctly and still produce wrong final answers. We introduce the \textbf{Novel Operator Test}, a benchmark that separates operator \emph{logic} from operator \emph{name}, enabling rigorous distinction between genuine reasoning and pattern retrieval. By evaluating Boolean operators under unfamiliar names across depths 1--10 on five models (up to 8,100 problems each), we demonstrate a \textbf{reasoning--output dissociation} that existing benchmarks cannot detect. At Claude Sonnet 4's depth 7, all 31 errors have verifiably correct reasoning yet wrong declared answers; 17/19 errors in mixed-operator chains exhibit the same pattern. This differs from CoT unfaithfulness \citep{turpin2024language}, where reasoning diverges from the actual process; here, reasoning \emph{is} correct yet the answer is wrong. The benchmark reveals two failure types: \emph{strategy failures} at depth 2, where models attempt terse retrieval (+62pp from scaffolding), and \emph{content failures} at depth 7, where models reason fully but err systematically (+8--30pp, 0/300 errors post-intervention). A Trojan operator (XOR's truth table under a novel name) confirms name alone does not gate reasoning ($p \geq 0.49$), while Llama's novelty gap widens to 28pp at depth 8--9 with the Trojan at 92--100\%, isolating genuine difficulty with novel logic from name unfamiliarity.
\end{abstract}

\section{Introduction}
\label{sec:intro}

Chain-of-thought (CoT) reasoning is widely used to improve and verify LLM outputs on logical tasks \citep{wei2022chain}. A natural assumption follows: if every reasoning step is correct, the final answer should be too. We show this assumption is false.

Using the \textbf{Novel Operator Test} (Boolean operators under unfamiliar names, evaluated across depths 1--10 on five models), we find that Claude Sonnet 4 produces \textbf{31 depth-7 errors where every CoT step is correct yet the declared answer is wrong}. We test nine operators: four standard (AND, OR, XOR, IF-THEN), four novel-named (BLIF, TARN, QUEX, DREM), and one Trojan (ZENT = XOR under a novel name), reporting four findings:

\begin{enumerate}
    \item \textbf{Reasoning--Output Dissociation}: All 31 depth-7 errors have correct CoT yet wrong answers; 17/19 mixed-operator chain errors show the same pattern. A self-correction probe (31/31 Claude, 30/31 GPT-4o) localizes the failure to autoregressive generation.
    \item \textbf{Two Failure Modes}: \emph{Strategy failures} at depth 2 (terse retrieval; +62pp from ETT scaffolding) and \emph{content failures} at depth 7 (full CoT but systematic errors; +8--30pp from ETT, 0/300 post-intervention).
    \item \textbf{Processing Strategy Divergence}: GPT-4o uses 141 tokens for standard operators vs.\ 343 for novel (2.4$\times$) at identical accuracy (99.5\%), with AND/OR answered in 1 token.
    \item \textbf{Model-Dependent Depth Profiles}: Llama's novelty gap widens from 11pp to 28pp at depth 8--9 while the Trojan stays at 92--100\%, isolating difficulty with novel logic from name unfamiliarity.
\end{enumerate}

\section{Related Work}
\label{sec:related}

\paragraph{Reasoning versus memorization.}
\citet{mccoy2024embers} show LLM behavior reflects statistical training patterns, with performance degrading on atypical examples. \citet{mirzadeh2024gsm} find that symbolic perturbations to GSM8K cause accuracy drops up to 65\%. \citet{zhao2024token} demonstrate that semantically irrelevant token changes cause large accuracy drops via token-level biases. Our work provides a complementary finding: accuracy on novel operators \emph{matches} familiar ones when given sufficient generation budget, but the behavioral strategy diverges sharply, a distinction that accuracy-based evaluation alone cannot detect.

\paragraph{Chain-of-thought and dual-process analogies.}
\citet{turpin2024language} show that chain-of-thought explanations can be unfaithful to the model's actual reasoning process; \citet{lanham2023measuring} further characterize when CoT influences model predictions. Our finding is complementary: we identify cases where CoT is \emph{faithful and correct} yet the final answer is wrong. \citet{hagendorff2023human} find LLMs exhibit human-like intuitive biases paralleling dual-process theory \citep{kahneman2011thinking}; our response length analysis provides consistent behavioral evidence.

\paragraph{Logical reasoning benchmarks.}
ProntoQA \citep{saparov2023language} tests syllogistic reasoning with fictional entities; LogicBench \citep{parmar2024logicbench} evaluates 25 reasoning patterns; ProofWriter \citep{tafjord2021proofwriter} generates proofs over natural language rules; FLD \citep{morishita2024fld} provides formal deduction benchmarks. These benchmarks isolate content novelty (new entities) while retaining standard logical connectives. We isolate \emph{rule-name} novelty.

\paragraph{Counterfactual and rule learning.}
\citet{wu2024reasoning} test LLMs on counterfactual task variants and find substantial drops; our novel operator \emph{names} avoid triggering preexisting associations. This complements the reversal curse \citep{berglund2023reversal}; our Trojan tests whether models can apply a known function under a new label. \citet{he2025idea} test rule discovery; we test rule \emph{application}. \citet{dziri2023faith} show transformers solve compositional tasks via linearized subgraph matching; \citet{dasgupta2024language} demonstrate content effects paralleling human cognitive biases.

\section{The Novel Operator Benchmark}
\label{sec:benchmark}

\subsection{Operator Design}
\label{sec:operators}

There are 16 binary Boolean functions. After removing six trivial functions (two constants, two projections, and two negated projections), ten non-trivial operators remain. We select nine, omitting NAND (to keep Group A at four operators) and converse nonimplication ($B \land \neg A$, which mirrors BLIF's structure).

\textbf{Group A (Standard)} contains AND, OR, XOR, and IF-THEN, which appear extensively in pretraining corpora.

\textbf{Group B (Novel-named)} contains four operators given unfamiliar names: BLIF (inhibition, $A \land \neg B$), TARN (NOR, $\neg(A \lor B)$), QUEX (converse implication, $B \to A$), and DREM (XNOR, $A \equiv B$). These sample the structural space along symmetry and sparsity (number of True rows).

\textbf{Group C (Trojan)} contains ZENT, which has XOR's exact truth table under a novel name. If ZENT performance matches XOR, the model applies operators from definitions regardless of name.

\begin{table}[t]
\centering
\small
\caption{Truth tables for all nine operators. Group B operators use novel names. ZENT (Group C) is XOR's truth table under a novel name, serving as the Trojan control.}
\label{tab:operators}
\begin{tabular}{llcccccc}
\toprule
Group & Name & TT & TF & FT & FF & Sym & \#T \\
\midrule
\multirow{4}{*}{A (Standard)} & AND & T & F & F & F & Y & 1 \\
& OR & T & T & T & F & Y & 3 \\
& XOR & F & T & T & F & Y & 2 \\
& IF-THEN & T & F & T & T & N & 3 \\
\midrule
\multirow{4}{*}{B (Novel)} & BLIF & F & T & F & F & N & 1 \\
& TARN & F & F & F & T & Y & 1 \\
& QUEX & T & T & F & T & N & 3 \\
& DREM & T & F & F & T & Y & 2 \\
\midrule
C (Trojan) & ZENT & F & T & T & F & Y & 2 \\
\bottomrule
\end{tabular}
\end{table}

\subsection{Problem Generation}

Each problem defines the relevant operator via truth table, specifies Boolean variable assignments, and asks the model to evaluate a left-associated chain expression.

\textbf{Depth 1} (single-step): evaluate $A \text{ OP } B$ given values of $A$ and $B$. \textbf{Depths 2--10}: evaluate left-associated homogeneous chains, e.g., $((A \text{ OP } B) \text{ OP } C) \text{ OP } D$ at depth 3. All variable assignments are sampled uniformly; we generate 50 instances per (operator, depth) condition.

The core benchmark contains 2{,}250 problems (9 operators $\times$ 5 depths $\times$ 50 instances) at depths 1--5, plus 2{,}250 extended problems at depths 6--10 (Section~\ref{sec:deep}), 1{,}200 problems testing representation format effects (Appendix~\ref{app:representation}), 1{,}200 testing ETT prompting at depths 2--5, and 600 testing ETT at depth 7 (Section~\ref{sec:ett}), plus 600 mixed-operator chain problems (Section~\ref{sec:discussion}), totaling \textbf{up to 8{,}100 problems per model} depending on applicable interventions (reasoning models o3-mini and QwQ-32B do not receive ETT or representation experiments; QwQ-32B is tested at depths 7 and 10 only in the extended experiment, yielding $\sim$3{,}150 problems).

\subsection{Prompting Interventions}

\textbf{Explicit Truth-Table Tracing (ETT).} A prompting intervention that forces step-by-step truth table lookup at each chain step, testing whether structured scaffolding improves performance on novel operators.

\section{Results}
\label{sec:results}

We evaluate five models: GPT-4o-2024-11-20, Claude Sonnet 4, Llama 3.1 70B Instruct, o3-mini-2025-01-31, and QwQ-32B, each completing up to 8{,}100 problems. Standard models use temperature 0.0 and max\_tokens$=$2048 (4{,}096 for depths 6--10); reasoning models use default temperature with max\_tokens$=$4{,}096 (8{,}192 for depths 6--10).

\subsection{Overview: Accuracy by Operator Group and Depth}
\label{sec:overview}

All models achieve $\geq$96\% at depth 1. At depth 5, GPT-4o and Claude Sonnet 4 achieve $\geq$96\% across all groups; Llama 3.1 70B shows a modest novelty gap. A notable exception: Llama's Group A accuracy dips to 91\% at depth 4, driven by IF-THEN at 68\%.

\begin{table}[t]
\centering
\scriptsize
\caption{Accuracy (\%) by operator group and chain depth. GPT-4o, Claude Sonnet 4, and QwQ-32B show no meaningful novelty gap at depth 5 ($|$gap$| \leq$6pp). Llama 3.1 70B shows an 11pp gap at depth 5, concentrated on structurally complex novel operators. $n=50$ instances per cell.}
\label{tab:main}
\setlength{\tabcolsep}{3.5pt}
\begin{tabular}{llccccccccc}
\toprule
& & \multicolumn{5}{c}{Accuracy (\%)} & \multicolumn{4}{c}{Novelty Gap (A$-$B, pp)} \\
\cmidrule(lr){3-7} \cmidrule(lr){8-11}
Model & Group & d=1 & d=2 & d=3 & d=4 & d=5 & d=2 & d=3 & d=4 & d=5 \\
\midrule
\multirow{3}{*}{GPT-4o} & A (Std) & 100 & 81.0 & 98.0 & 97.5 & 99.5 & & & & \\
& B (Novel) & 100 & 69.0 & 99.0 & 99.5 & 99.5 & 12.0 & $-$1.0 & $-$2.0 & 0 \\
& C (Trojan) & 100 & 82.0 & 98.0 & 100 & 100 & & & & \\
\midrule
\multirow{3}{*}{Claude Sonnet 4} & A (Std) & 100 & 95.0 & 92.5 & 96.5 & 97.5 & & & & \\
& B (Novel) & 100 & 93.0 & 94.0 & 94.0 & 98.5 & 2.0 & $-$1.5 & 2.5 & $-$1.0 \\
& C (Trojan) & 100 & 100 & 94.0 & 94.0 & 96.0 & & & & \\
\midrule
\multirow{3}{*}{Llama 3.1 70B} & A (Std) & 100 & 98.5 & 100 & 91.0 & 98.5 & & & & \\
& B (Novel) & 100 & 91.5 & 91.5 & 93.5 & 87.5 & 7.0 & 8.5 & $-$2.5 & \textbf{11.0} \\
& C (Trojan) & 100 & 88.0 & 92.0 & 96.0 & 96.0 & & & & \\
\midrule
\multirow{3}{*}{o3-mini} & A (Std) & 100 & 100 & 100 & 100 & 100 & & & & \\
& B (Novel) & 100 & 100 & 100 & 100 & 100 & 0 & 0 & 0 & 0 \\
& C (Trojan) & 100 & 100 & 100 & 100 & 98.0 & & & & \\
\midrule
\multirow{3}{*}{QwQ-32B} & A (Std) & 100 & 99.5 & 100 & 94.5 & 100 & & & & \\
& B (Novel) & 99.0 & 95.5 & 93.5 & 93.0 & 94.0 & 4.0 & 6.5 & 1.5 & \textbf{6.0} \\
& C (Trojan) & 96.0 & 74.0 & 82.0 & 100 & 100 & & & & \\
\bottomrule
\end{tabular}
\end{table}

The novelty gap (Group A $-$ Group B) is negligible for GPT-4o (0pp at depth 5) and Claude Sonnet 4 ($-$1pp), but present for Llama (11pp, $p < 0.01$, Welch's $t$-test) and moderate for QwQ-32B (6pp). Per-operator analysis (Appendix~\ref{app:per_operator_d5}) shows Llama's gap concentrates on QUEX (78\%) and DREM (86\%), confirming that structural properties influence difficulty independently of familiarity.

\begin{figure}[t]
\centering
\includegraphics[width=\linewidth]{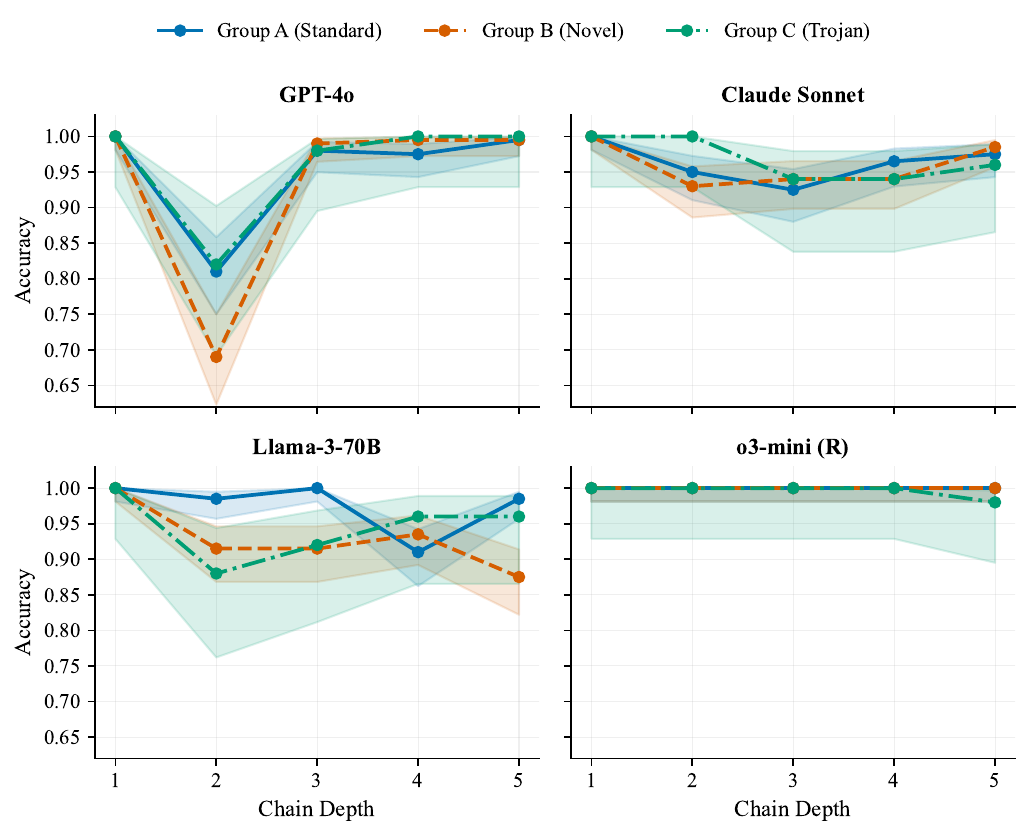}
\caption{Accuracy by operator group and chain depth (1--5) across five models. The depth-4 data point fills the recovery trajectory between the depth-2 dip and depth-5 ceiling. Llama shows a notable non-monotonic dip at depth 4 (91\% Group A), driven by IF-THEN (68\%). All conditions use $n=50$ instances; shaded regions show 95\% binomial CIs.}
\label{fig:accuracy_depth}
\end{figure}

\subsection{The Trojan Operator: XOR vs.\ ZENT}
\label{sec:trojan}

ZENT has XOR's exact truth table under a novel name, differing only in the operator name ($n=50$ per cell, independently sampled).

\begin{table}[t]
\centering
\scriptsize
\caption{XOR vs.\ ZENT accuracy (\%) by depth ($n=50$ per cell). No model shows a significant name effect at depth 5 (Fisher's exact $p \geq 0.49$). 95\% binomial CIs shown for depth 5; Fisher's $p$ in rightmost column.}
\label{tab:trojan}
\setlength{\tabcolsep}{3pt}
\begin{tabular}{lcccccccccccc}
\toprule
& \multicolumn{2}{c}{d=1} & \multicolumn{2}{c}{d=2} & \multicolumn{2}{c}{d=3} & \multicolumn{2}{c}{d=4} & \multicolumn{2}{c}{d=5} & \\
\cmidrule(lr){2-3} \cmidrule(lr){4-5} \cmidrule(lr){6-7} \cmidrule(lr){8-9} \cmidrule(lr){10-11}
Model & XOR & ZENT & XOR & ZENT & XOR & ZENT & XOR & ZENT & XOR & ZENT & $p$ \\
\midrule
GPT-4o & 100 & 100 & 62 & 82 & 92 & 98 & 90 & 100 & 98{\scriptsize$\pm$4} & 100{\scriptsize$\pm$0} & 1.0 \\
Claude Sonnet 4 & 100 & 100 & 98 & 100 & 86 & 94 & 100 & 94 & 100{\scriptsize$\pm$0} & 96{\scriptsize$\pm$5} & 0.49 \\
Llama 3.1 70B & 100 & 100 & 100 & 88 & 100 & 92 & 96 & 96 & 98{\scriptsize$\pm$4} & 96{\scriptsize$\pm$5} & 1.0 \\
\midrule
o3-mini & 100 & 100 & 100 & 100 & 100 & 100 & 100 & 100 & 100{\scriptsize$\pm$0} & 98{\scriptsize$\pm$4} & 1.0 \\
\midrule
QwQ-32B & 100 & 96 & 98 & 74 & 100 & 82 & 92 & 100 & 100{\scriptsize$\pm$0} & 100{\scriptsize$\pm$0} & 1.0 \\
\bottomrule
\end{tabular}
\end{table}

Table~\ref{tab:trojan} presents the central result: at depth 5, \textbf{no model shows a significant XOR--ZENT gap} (Fisher's exact $p \geq 0.49$ for all). At depth 2, GPT-4o shows a reversal: ZENT (82\%) outperforms XOR (62\%), suggesting name familiarity can \emph{hurt} at the strategy--reasoning transition.

\subsection{The Depth-2 Reasoning Mode Transition}
\label{sec:depth2}

\begin{figure}[t]
\centering
\includegraphics[width=\linewidth]{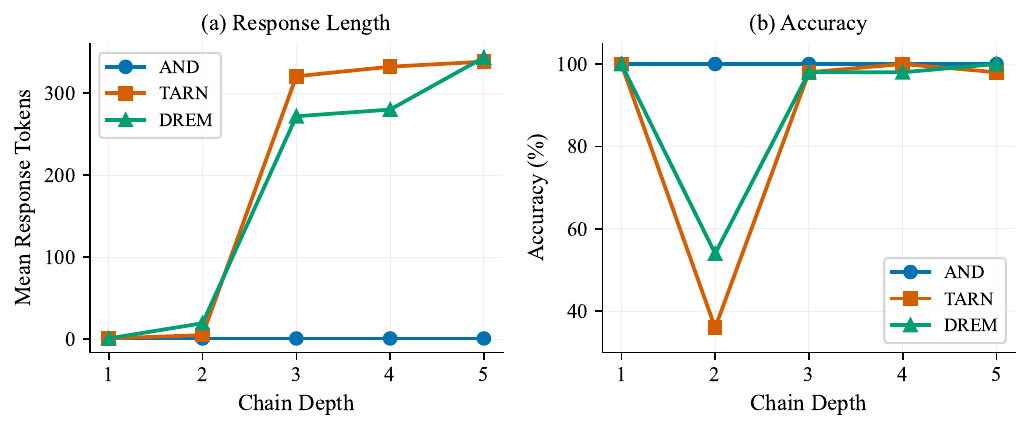}
\caption{GPT-4o response length and accuracy by chain depth for selected operators. (a) AND stays at 1 token across all depths (pure retrieval); TARN and DREM jump from $<$20 tokens at depth 2 to 300+ tokens at depth 3, marking the transition from retrieval to chain-of-thought reasoning. (b) Accuracy dips at depth 2 (the transition point) and recovers at depth 3+ once models adopt explicit reasoning.}
\label{fig:depth_transition}
\end{figure}

GPT-4o's accuracy follows a non-monotonic trajectory across depths on most operators: 100\% at depth 1, a sharp dip at depth 2, then recovery through depths 3--5. Response length data reveals the mechanism (Figure~\ref{fig:depth_transition}). At depth 1, GPT-4o answers all operators in 1 token (retrieval suffices for single-step lookup). At depth 2, the model still attempts brief responses: XOR averages 1 token (62\% accuracy), TARN averages 5 tokens (36\%), DREM averages 20 tokens (54\%). By depth 3, the model commits to full chain-of-thought reasoning: TARN jumps to 321 tokens (98\% accuracy), DREM to 272 tokens (98\%).

The depth-2 dip marks a \textbf{reasoning mode transition}: models attempt retrieval-style responses for multi-step composition, and these short responses fail. XOR is particularly informative: GPT-4o answers all 50 depth-2 instances in exactly 1 token, achieving only 62\%. The transition is operator-dependent: BLIF achieves 96\% at depth 2 (one True row yields predictable chains), while operators with balanced truth tables (TARN: 36\%, DREM: 54\%) suffer sharply. This is precisely where ETT scaffolding has its largest effect (Section~\ref{sec:ett}).

\subsection{Behavioral Strategy Divergence: Response Length Analysis}
\label{sec:response_length}

We measure \textbf{response length} (completion tokens) as a behavioral proxy for whether models answer via terse retrieval or explicit chain-of-thought.

\begin{table}[t]
\centering
\small
\caption{Mean response length (completion tokens $\pm$ std.\ dev.) at depth 5 by operator group ($n=200$ for A/B, $n=50$ for C). GPT-4o's high Group A variance reflects the bimodal distribution: AND/OR produce 1 token while XOR/IF-THEN produce $\sim$300.}
\label{tab:response_length}
\begin{tabular}{lcccccc}
\toprule
& \multicolumn{2}{c}{Group A (Std)} & \multicolumn{2}{c}{Group B (Novel)} & \multicolumn{2}{c}{Group C (Trojan)} \\
\cmidrule(lr){2-3} \cmidrule(lr){4-5} \cmidrule(lr){6-7}
Model & Tokens & Acc & Tokens & Acc & Tokens & Acc \\
\midrule
GPT-4o & 141{\scriptsize$\pm$181} & 99.5\% & 343{\scriptsize$\pm$67} & 99.5\% & 357{\scriptsize$\pm$35} & 100\% \\
Claude Sonnet 4 & 199{\scriptsize$\pm$54} & 97.5\% & 272{\scriptsize$\pm$15} & 98.5\% & 279{\scriptsize$\pm$16} & 96\% \\
Llama 3.1 70B & 103{\scriptsize$\pm$180} & 98.5\% & 124{\scriptsize$\pm$168} & 87.5\% & 241{\scriptsize$\pm$117} & 96\% \\
o3-mini & 227{\scriptsize$\pm$176} & 100\% & 630{\scriptsize$\pm$266} & 100\% & 375{\scriptsize$\pm$118} & 98\% \\
\bottomrule
\end{tabular}
\end{table}

Table~\ref{tab:response_length} shows that GPT-4o uses 141 mean tokens for Group A and 343 for Group B at depth 5: a 2.4$\times$ processing cost difference at identical accuracy (both 99.5\%). The divergence is sharpest at the operator level: GPT-4o answers AND and OR in exactly \textbf{1 token} (pure retrieval, the model outputs ``True'' or ``False'' without reasoning) but requires 335+ tokens for every novel and Trojan operator (Group B mean: 343, Group C: 357). XOR occupies an intermediate position (187 tokens), consistent with partial memorization.

Claude and o3-mini show consistent but smaller divergences (Table~\ref{tab:response_length}). Llama is the only model where longer novel-operator processing does \emph{not} fully compensate: Group B accuracy (87.5\%) lags Group A (98.5\%) despite additional reasoning.

\begin{figure}[t]
\centering
\includegraphics[width=\linewidth]{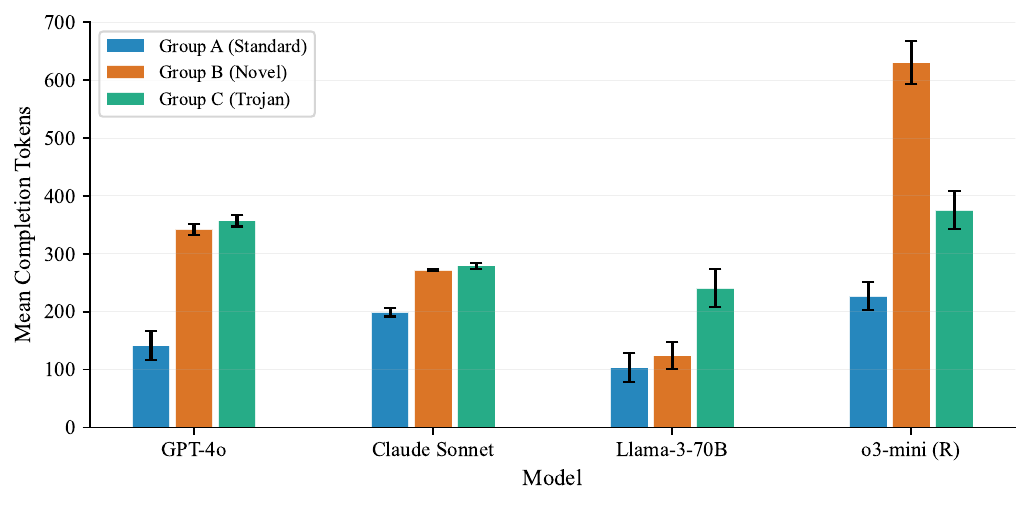}
\caption{Mean completion tokens at depth 5 by operator group across four of five models (QwQ-32B excluded; response length data not collected). GPT-4o shows the starkest divergence between standard (141 tokens) and novel (343 tokens) groups. Error bars show 95\% CIs.}
\label{fig:response_length}
\end{figure}

\subsection{Error Analysis: Standard Operator Interference}
\label{sec:sois}

Error analysis reveals that models substitute \emph{specific} familiar operators rather than producing random errors. At depth 2, all 14 of Claude's DREM errors produce AND-chain output ($p < 10^{-6}$), and all 6 of Llama's match OR ($p = 0.001$). At depths 3--5, the substitution target shifts to XOR (structurally similar to XNOR), paralleling the reasoning mode transition from simple-operator defaults to structurally-similar-operator confusion (full analysis in Appendix~\ref{app:error_analysis}).

\subsection{ETT Prompting}
\label{sec:ett}

We test Explicit Truth-Table Tracing (ETT) on Group B operators at depths 2, 3, and 5 for the three standard models. ETT forces the model to explicitly name each intermediate step, identify the relevant truth table row, and state the result before proceeding. We exclude o3-mini because it achieves 100\% across all conditions, making ETT effects undetectable.

\begin{table}[t]
\centering
\scriptsize
\caption{ETT improvement (pp) on Group B operators ($n=50$ per cell). ETT has its largest effect at depth 2, where it bridges the reasoning mode transition (Section~\ref{sec:depth2}). At depths 3--5, ETT helps Llama and Claude but has no effect on GPT-4o (already at ceiling). The ETT experiment uses paired within-experiment baselines with separate variable assignments (seed 44) from the core experiment (seed 42).}
\label{tab:ett}
\setlength{\tabcolsep}{2.8pt}
\begin{tabular}{lcccccccccccc}
\toprule
& \multicolumn{4}{c}{Depth 2} & \multicolumn{4}{c}{Depth 3} & \multicolumn{4}{c}{Depth 5} \\
\cmidrule(lr){2-5} \cmidrule(lr){6-9} \cmidrule(lr){10-13}
Model & BLIF & TARN & QUEX & DREM & BLIF & TARN & QUEX & DREM & BLIF & TARN & QUEX & DREM \\
\midrule
GPT-4o & +10 & \textbf{+62} & +6 & +26 & 0 & 0 & 0 & 0 & 0 & 0 & 0 & 0 \\
Claude Sonnet 4 & 0 & 0 & 0 & +30 & 0 & +14 & +8 & +12 & 0 & 0 & +4 & +12 \\
Llama 3.1 70B & +10 & +6 & +22 & +10 & 0 & +12 & +10 & +16 & +22 & +2 & +16 & +16 \\
\bottomrule
\end{tabular}
\end{table}

Table~\ref{tab:ett} shows a depth-dependent pattern (improvement values against paired baselines, seed 44). At \textbf{depth 2}, ETT has its largest effects: GPT-4o improves +62pp on TARN and +26pp on DREM by forcing step-by-step reasoning. At \textbf{depths 3--5}, GPT-4o shows zero improvement (already at ceiling), while Claude and Llama benefit moderately. DREM (XNOR) shows the most consistent benefit across all models, consistent with ETT helping avoid XOR/XNOR confusion.

\paragraph{ETT at depth 7: addressing content failures.}
We extend ETT to Claude's depth-7 dip, where the model already produces full chain-of-thought ($\sim$340 tokens). Table~\ref{tab:ett_d7} shows ETT achieves \textbf{100\% on all six operators} (0/300 errors vs.\ 34/300 baseline; aggregate $p < 10^{-6}$). All 34 baseline errors occur on True-ground-truth instances. Since the depth-7 ground truth base rate is 52\% True across operators (and 38--72\% for the four affected operators individually), this directional bias is highly significant ($p < 10^{-6}$, binomial test against base rate), suggesting the model defaults to a ``False'' prior when uncertainty accumulates over long chains. ETT adds $\sim$140 tokens of structured scaffolding, demonstrating that Claude \emph{can} compute correctly at depth 7 when guided through explicit truth-table lookups.

\begin{table}[t]
\centering
\scriptsize
\caption{ETT improvement at depth 7 for Claude Sonnet 4 ($n=50$ per cell, seed 142). ETT achieves 100\% on all operators. The four operators with baseline errors show +8 to +30pp improvement; controls (BLIF, XOR) are unaffected. All 34 baseline errors answer False on True-ground-truth instances. Note: the deep experiment (Section~\ref{sec:deep}, different seed) produces 31 errors with the same directional pattern.}
\label{tab:ett_d7}
\setlength{\tabcolsep}{4pt}
\begin{tabular}{lcccccc}
\toprule
& IF-THEN & ZENT & TARN & DREM & BLIF & XOR \\
\midrule
Baseline (\%) & 70 & 82 & 88 & 92 & 100 & 100 \\
ETT (\%) & \textbf{100} & \textbf{100} & \textbf{100} & \textbf{100} & 100 & 100 \\
$\Delta$ (pp) & +30 & +18 & +12 & +8 & 0 & 0 \\
Fisher's $p$ & ${<}0.0001$ & $0.003$ & $0.027$ & $0.12$ & --- & --- \\
Baseline errors & 15 & 9 & 6 & 4 & 0 & 0 \\
\quad on GT=True & 15 & 9 & 6 & 4 & --- & --- \\
\quad on GT=False & 0 & 0 & 0 & 0 & --- & --- \\
\bottomrule
\end{tabular}
\end{table}

\paragraph{Programmatic error chain verification.}
Programmatic verification of all 34 depth-7 baseline errors (Appendix~\ref{app:depth7_examples}) confirms that \textbf{every error chain correctly computes ``True'' at the final step, then declares ``False'' as the answer} (IF-THEN: 15/15, ZENT: 9/9, TARN: 6/6, DREM: 4/4), a reasoning--output dissociation where the failure lies in the answer declaration, not the reasoning process.

\paragraph{Self-correction probe.}
We present each of Claude's 31 error chains to both Claude and GPT-4o in \emph{new conversations} and ask what value was computed in the last step. Claude self-corrects on all 31 (100\%) and GPT-4o on 30/31 (97\%), confirming both models can extract the correct answer from the reasoning trace, localizing the dissociation to autoregressive generation rather than reasoning failure.

\paragraph{Temperature stability analysis.}
We re-run Claude at depth 7 across $T \in \{0.0, 0.5, 1.0\}$ (2{,}700 total probes). The dissociation rate is \textbf{temperature-stable}: 57/900 at $T{=}0.0$, 53/900 at $T{=}0.5$, 51/900 at $T{=}1.0$. For the 31 originally-wrong instances, 46\% remain dissociated at $T{=}0.0$, confirming a structural rather than stochastic property.

Code-format operator definitions match or exceed truth tables at depth 2 for all models (Appendix~\ref{app:representation}). Restrictive token limits (max\_tokens$=$256) create phantom accuracy collapses of 30--54pp by truncating verbose reasoning on novel operators (Appendix~\ref{app:confound}).

\subsection{Extended Depth Analysis (d=6--10)}
\label{sec:deep}

To test whether the novelty gap emerges at greater chain depths, we extend the benchmark to depths 6--10, generating up to 2{,}250 additional problems per model (9 operators $\times$ 5 depths $\times$ 50 instances; QwQ-32B tested at d=7 and d=10 only due to inference cost). Table~\ref{tab:deep} summarizes group-level accuracy for all five models.

\begin{table}[t]
\centering
\scriptsize
\caption{Group-level accuracy (\%) at depths 6--10 ($n=200$ per group cell, $n=50$ for Trojan). GPT-4o and o3-mini maintain near-perfect accuracy. Claude shows a depth-7 dip with recovery. Llama's novelty gap \emph{widens} from 11pp at depth 5 to 28pp at depth 8--9. QwQ-32B tested at d=7 and d=10 only (---); its novelty gap vanishes by d=10.}
\label{tab:deep}
\setlength{\tabcolsep}{3pt}
\begin{tabular}{llccccc}
\toprule
Model & Group & d=6 & d=7 & d=8 & d=9 & d=10 \\
\midrule
\multirow{3}{*}{GPT-4o} & A (Std) & 99.5 & 100 & 99.0 & 100 & 100 \\
& B (Novel) & 99.0 & 100 & 99.5 & 99.0 & 99.0 \\
& C (Trojan) & 100 & 100 & 98.0 & 100 & 100 \\
\midrule
\multirow{3}{*}{Claude Sonnet 4} & A (Std) & 99.5 & 96.0 & 97.5 & 99.0 & 99.5 \\
& B (Novel) & 94.5 & 93.5 & 100 & 99.5 & 99.0 \\
& C (Trojan) & 96.0 & \textbf{80.0} & 100 & 100 & 100 \\
\midrule
\multirow{3}{*}{Llama 3.1 70B} & A (Std) & 97.5 & 98.5 & 94.5 & 94.0 & 94.0 \\
& B (Novel) & 89.0 & 84.0 & \textbf{66.5} & \textbf{66.0} & 76.5 \\
& C (Trojan) & 94.0 & 98.0 & 96.0 & 92.0 & 100 \\
\midrule
\multirow{3}{*}{o3-mini} & A (Std) & 100 & 100 & 100 & 100 & 100 \\
& B (Novel) & 100 & 100 & 100 & 100 & 99.5 \\
& C (Trojan) & 100 & 100 & 100 & 100 & 100 \\
\midrule
\multirow{3}{*}{QwQ-32B} & A (Std) & --- & 93.5 & --- & --- & 88.5 \\
& B (Novel) & --- & 92.5 & --- & --- & 89.5 \\
& C (Trojan) & --- & 86.0 & --- & --- & 90.0 \\
\bottomrule
\end{tabular}
\end{table}

GPT-4o and o3-mini maintain $\geq$99\% through depth 10. \textbf{Llama 3.1 70B} reveals a \textbf{widening novelty gap}: 11pp at depth 5 to 28pp at depth 8--9, with QUEX collapsing to 40\% at d=9. The Trojan ZENT remains at 92--100\%, providing strong evidence that the gap reflects difficulty with novel operator \emph{logic}, not name unfamiliarity. Response length confirms these are strategy failures: QUEX errors at d=9 average 2 tokens vs.\ 340 for correct answers.

\begin{figure}[t]
\centering
\includegraphics[width=\linewidth]{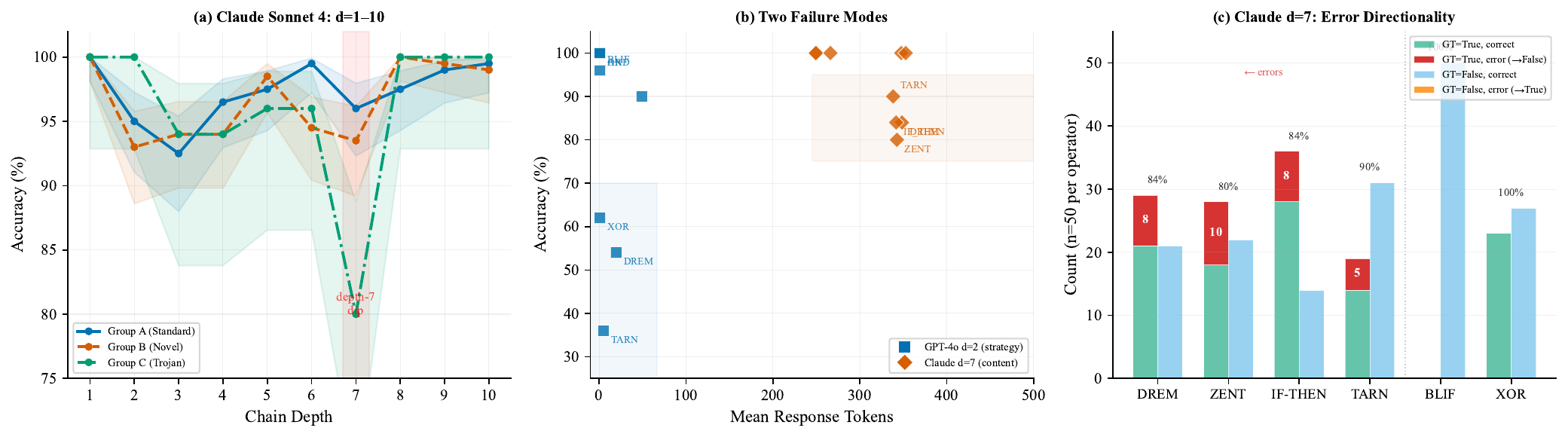}
\caption{Two failure modes in LLM logical reasoning. (a) Claude Sonnet 4 accuracy across depths 1--10 by operator group, showing the depth-7 dip with full recovery by depth 8. (b) Strategy failures (GPT-4o d=2, blue squares): low token counts, low accuracy; models attempt retrieval over chain-of-thought. Content failures (Claude d=7, orange diamonds): high token counts ($\sim$350), moderate accuracy; full reasoning produces wrong answers. (c) Error directionality at Claude's depth 7: all 31 errors (red) occur on True-ground-truth instances; zero errors on False-ground-truth instances. BLIF and XOR (100\% accuracy) serve as controls.}
\label{fig:failure_modes}
\end{figure}

\textbf{Claude Sonnet 4} reveals a non-monotonic \textbf{depth-7 dip} with full recovery by depth 8 (Figure~\ref{fig:failure_modes}a). Unlike Llama's strategy failures, Claude's 31 depth-7 errors (from this seed; the ETT experiment with seed 142 produces 34 with the same pattern) involve full CoT ($\sim$348 tokens) yet all answer False on True-ground-truth instances (Figure~\ref{fig:failure_modes}b--c), the \textbf{content failure} pattern from Section~\ref{sec:ett}. We hypothesize depth 7 sits at a critical threshold where chain length first exceeds the model's reliable context for answer extraction: depths 1--6 are short enough for robust summarization, while at depth 8+ the model may adopt a more cautious strategy that restores accuracy.

\section{Discussion}
\label{sec:discussion}

\paragraph{Reasoning--output dissociation.}
Our dissociation is the converse of \citet{turpin2024language}'s unfaithful CoT: here, CoT is faithful and correct yet produces the wrong answer, a failure of \emph{output generation} rather than \emph{explanation}. The self-correction probe (31/31) and temperature stability (stable across $T{=}0.0$--$1.0$) localize this to a structural property of autoregressive generation.

\paragraph{Differential internalization.}
The 2.4$\times$ token divergence at equivalent accuracy reveals familiar operators are \emph{internalized} (1-token retrieval) while novel operators require explicit reasoning. The two failure modes (strategy failures at depth 2, +62pp from ETT; content failures at depth 7, +8--30pp) show distinct remediation profiles.

\paragraph{Mixed-operator chains.}
To test whether the dissociation is an artifact of homogeneous chains, we evaluate mixed-operator chains at depths 3, 5, and 7 (50 instances each) under two conditions: \emph{all-novel} (cycling BLIF$\to$TARN$\to$QUEX$\to$DREM) and \emph{novel-standard alternating}. Claude Sonnet 4 shows \textbf{17/19 errors (89\%) are dissociations} across both conditions, confirming the phenomenon is not an artifact of operator repetition. Novel-standard alternating chains drop to 90\% at depth 7 as compositional complexity increases; GPT-4o achieves 82--100\% with errors concentrated at depth 7.

\paragraph{Limitations.}
Response length is a behavioral proxy; mechanistic probing would confirm internal route distinctions. Frontier models maintain near-ceiling accuracy at depth 10, so non-Boolean logic (e.g., first-order or probabilistic reasoning) may be needed to stress-test whether the dissociation generalizes to richer domains. Sample size ($n{=}50$) suffices for the large effects reported but limits detection of small per-operator differences; the DREM ETT improvement at depth 7 (+8pp, $p=0.12$) does not reach significance individually, though it contributes to the significant aggregate effect.

\section{Conclusion}
\label{sec:conclusion}

We introduced the Novel Operator Test, a benchmark that disentangles reasoning from retrieval by separating operator logic from operator name, and identified a \emph{reasoning--output dissociation}: correct CoT yet wrong answers (31/31 homogeneous, 17/19 mixed chains). This failure mode is invisible to benchmarks that use familiar operators, because retrieval and reasoning are confounded. Two distinct failure modes (strategy failures at depth 2 and content failures at depth 7) respond differently to structured scaffolding. Verifying chain-of-thought correctness is insufficient to guarantee answer correctness.

\bibliographystyle{iclr2026_conference}
\bibliography{references}

\appendix
\section{Prompt Templates}
\label{app:prompts}

The standard prompt provides the operator truth table, variable assignments, and asks for the answer (``True'' or ``False''). The ETT prompt additionally forces step-by-step truth table lookup:

\subsection{ETT Prompt (Depth 2, Novel Operator)}

\begin{verbatim}
You are given the following logical operator(s):

BLIF is defined by the following truth table:
  A=True,  B=True  -> False
  A=True,  B=False -> True
  A=False, B=True  -> False
  A=False, B=False -> False

Question: If A=True, B=False, C=True,
  what is (A BLIF B) BLIF C?

Solve by explicitly looking up each step
  in the truth table:
Step 1: Identify the innermost expression:
  A BLIF B, where A=True, B=False.
  Look up row (True, False) in the BLIF
  truth table. Result = ___
Step 2: Let R1 = your result from Step 1.
  Now evaluate R1 BLIF C, where R1=___,
  C=True. Look up row (R1, True) in the
  BLIF truth table. Result = ___

Final answer (True or False):
\end{verbatim}

\section{Representation Format Results}
\label{app:representation}

\begin{table}[h]
\centering
\small
\caption{Accuracy (\%) by representation format for Group B operators ($n=200$ per cell, aggregated across 4 operators). Code format matches or exceeds truth tables at depth 2 for all models.}
\label{tab:representation}
\begin{tabular}{lcccccc}
\toprule
& \multicolumn{3}{c}{Depth 1} & \multicolumn{3}{c}{Depth 2} \\
\cmidrule(lr){2-4} \cmidrule(lr){5-7}
Model & Table & Code & NL & Table & Code & NL \\
\midrule
GPT-4o & \textbf{100} & 72 & 89.5 & 72.5 & \textbf{93.5} & 76 \\
Claude Sonnet 4 & 100 & 100 & 100 & 92.5 & \textbf{100} & 90.5 \\
Llama 3.1 70B & 100 & 88 & 100 & 86.5 & \textbf{98} & \textbf{98.5} \\
\bottomrule
\end{tabular}
\end{table}

\section{Per-Operator Accuracy at Depth 5}
\label{app:per_operator_d5}

\begin{table}[h]
\centering
\small
\caption{Accuracy (\%) per operator at depth 5 ($n=50$ per cell). GPT-4o, Claude, and o3-mini achieve $\geq$90\% on all operators. Llama shows difficulty on QUEX (78\%) and DREM (86\%); QwQ-32B shows difficulty specifically on DREM (78\%).}
\label{tab:peroperator}
\begin{tabular}{llccccccc}
\toprule
Grp & Operator & Sym & \#T & GPT-4o & Claude & Llama & o3-mini & QwQ \\
\midrule
\multirow{4}{*}{A} & AND & Y & 1 & 100 & 100 & 100 & 100 & 100 \\
& OR & Y & 3 & 100 & 100 & 100 & 100 & 100 \\
& XOR & Y & 2 & 98 & 100 & 98 & 100 & 100 \\
& IF-THEN & N & 3 & 100 & 90 & 96 & 100 & 100 \\
\midrule
\multirow{4}{*}{B} & BLIF & N & 1 & 100 & 100 & 94 & 100 & 100 \\
& TARN & Y & 1 & 98 & 98 & 92 & 100 & 98 \\
& QUEX & N & 3 & 100 & 96 & \textbf{78} & 100 & 100 \\
& DREM & Y & 2 & 100 & 100 & \textbf{86} & 100 & \textbf{78} \\
\midrule
C & ZENT & Y & 2 & 100 & 96 & 96 & 98 & 100 \\
\bottomrule
\end{tabular}
\end{table}

\section{Token Limits as a Methodological Confound}
\label{app:confound}

With max\_tokens$=$256, we observed apparent accuracy collapses of 30--54pp between XOR and ZENT at depth 5 (e.g., GPT-4o: XOR 98\%, ZENT 44\%). These were \textbf{truncation artifacts}: novel operators elicit verbose responses exceeding 256 tokens. Re-running with max\_tokens$=$2048 eliminated the artifact (ZENT: 44\% $\to$ 100\%). \textbf{Restrictive token limits interact with operator familiarity to create phantom performance gaps.}

\section{GPT-4o Full Per-Operator Results}
\label{app:per_operator}

\begin{table}[h]
\centering
\small
\caption{Full per-operator accuracy (\%) across all depths for GPT-4o. Non-monotonic depth-2 dip and mean response tokens at depth 2 illustrate the reasoning mode transition. Depth 4 fills the recovery trajectory between depth 3 and 5.}
\label{tab:gpt4o_full}
\begin{tabular}{llcccccc}
\toprule
Group & Operator & d=1 & d=2 & d=3 & d=4 & d=5 & d=2 tokens \\
\midrule
\multirow{4}{*}{A} & AND & 100 & 100 & 100 & 100 & 100 & 1 \\
& OR & 100 & 100 & 100 & 100 & 100 & 1 \\
& XOR & 100 & 62 & 92 & 90 & 98 & 1 \\
& IF-THEN & 100 & 62 & 100 & 100 & 100 & 1 \\
\midrule
\multirow{4}{*}{B} & BLIF & 100 & 96 & 100 & 100 & 100 & 1 \\
& TARN & 100 & 36 & 98 & 100 & 98 & 5 \\
& QUEX & 100 & 90 & 100 & 100 & 100 & 49 \\
& DREM & 100 & 54 & 98 & 98 & 100 & 20 \\
\midrule
C & ZENT & 100 & 82 & 98 & 100 & 100 & 52 \\
\bottomrule
\end{tabular}
\end{table}

\section{Prompt-Compliance Control Experiment}
\label{app:show_work}

A potential confound for the response length analysis is that the standard prompt instructs ``Answer with only True or False,'' which may directly cause GPT-4o's 1-token responses on familiar operators. To test this, we run a control experiment replacing the instruction with ``Show your step-by-step reasoning, then state your final answer as FINAL ANSWER: True or FINAL ANSWER: False'' ($n=30$ per condition, 480 problems total).

\begin{table}[h]
\centering
\small
\caption{Standard vs.\ ``show your work'' prompts for GPT-4o ($n=30$ per cell). Show-work prompt forces verbose reasoning (342--770 tokens for AND/OR) and eliminates depth-2 strategy failures (TARN: 33\%$\to$100\%).}
\label{tab:show_work}
\begin{tabular}{llcccc}
\toprule
& & \multicolumn{2}{c}{Accuracy (\%)} & \multicolumn{2}{c}{Mean tokens} \\
\cmidrule(lr){3-4} \cmidrule(lr){5-6}
Operator & Depth & Standard & Show-work & Standard & Show-work \\
\midrule
AND & 2 & 100 & 100 & 1 & 361 \\
AND & 5 & 100 & 100 & 1 & 770 \\
OR & 2 & 100 & 100 & 1 & 343 \\
OR & 5 & 100 & 100 & 1 & 749 \\
BLIF & 2 & 87 & \textbf{100} & 1 & 475 \\
BLIF & 5 & 100 & 100 & 320 & 1079 \\
TARN & 2 & \textbf{33} & \textbf{100} & 8 & 502 \\
TARN & 5 & 100 & 100 & 349 & 894 \\
\bottomrule
\end{tabular}
\end{table}

The 1-token standard responses reflect prompt compliance, but the critical asymmetry remains: \textbf{AND/OR are correct in 1 token while TARN achieves only 33\%}, supporting \emph{differential internalization}.

\section{Error Analysis: Standard Operator Interference (Full)}
\label{app:error_analysis}

We check whether each wrong answer matches a specific standard operator on the same inputs. DREM errors show operator-specific substitution: at depth 2, all 14 of Claude's errors produce AND-chain output ($p < 10^{-6}$), all 6 of Llama's match OR ($p = 0.001$). At depths 3--5, the target shifts to XOR.

\section{Depth-7 Error Examples and Programmatic Verification}
\label{app:depth7_examples}

All 31 depth-7 errors have correct CoT through all 7 steps but declare the wrong answer. Two examples:

\paragraph{Example 1: IF-THEN at depth 7.}
\begin{small}
\begin{verbatim}
Variables: A=True, B=True, C=False, D=False, E=True, F=True, G=False, H=False
1) A IF-THEN B = True IF-THEN True = True
2) (A IF-THEN B) IF-THEN C = True IF-THEN False = False
3) (...) IF-THEN D = False IF-THEN False = True
4) (...) IF-THEN E = True IF-THEN True = True
5) (...) IF-THEN F = True IF-THEN True = True
6) (...) IF-THEN G = True IF-THEN False = False
7) (...) IF-THEN H = False IF-THEN False = True    <-- correct

False                                               <-- declared answer
\end{verbatim}
\end{small}
\paragraph{Example 2: ZENT (Trojan = XOR) at depth 7.}
\begin{small}
\begin{verbatim}
Variables: A=True, B=True, C=False, D=False, E=False, F=True, G=True, H=True
1) A ZENT B = True ZENT True = False
2) (...) ZENT C = False ZENT False = False
3) (...) ZENT D = False ZENT False = False
4) (...) ZENT E = False ZENT False = False
5) (...) ZENT F = False ZENT True = True
6) (...) ZENT G = True ZENT True = False
7) (...) ZENT H = False ZENT True = True            <-- correct

False                                               <-- declared answer
\end{verbatim}
\end{small}
This pattern holds for all 31 errors (IF-THEN: 8, ZENT: 10, DREM: 8, TARN: 5). All have ground truth True.

\end{document}